\documentclass[11pt]{article}

\usepackage[preprint]{acl}

\usepackage{times}
\usepackage{latexsym}

\usepackage[T1]{fontenc}
\usepackage[utf8]{inputenc}

\usepackage{microtype}

\usepackage{inconsolata}
\usepackage{url}
\usepackage{graphicx}
\usepackage{listings}
\usepackage{svg}
\usepackage{multirow}
\usepackage{enumitem}
\usepackage{makecell}
\usepackage{tabularx}
\newcommand{\dataset}[0]{M{\sc ulti}C{\sc om}}

\title{Do LLMs exhibit the same commonsense capabilities across languages?}

\author{
  \textbf{Iván Martínez-Murillo\textsuperscript{1}},
  \textbf{Elena Lloret\textsuperscript{1}},
  \textbf{Paloma Moreda\textsuperscript{1}},\\
  \textbf{Albert Gatt\textsuperscript{2}},
  ,
\\
  \textsuperscript{1}University of Alicante, Carr. de San Vicente del Raspeig, s/n, \\San Vicente del Raspeig, Alicante, Spain, 03690\\
  \textsuperscript{2}University of Utrecht, Department of Information and Computing Sciences, \\Princetonplein 5, 3584 Utrecht, The Netherlands
\\
}
\begin{document}
\maketitle
\begin{abstract}
This paper explores the multilingual commonsense generation abilities of Large Language Models (LLMs). To facilitate this investigation, we introduce MULTICOM, a novel benchmark that extends the COCOTEROS dataset to four languages: English, Spanish, Dutch, and Valencian. The task involves generating a commonsensical sentence that includes a given triplet of words. We evaluate a range of open-source LLMs, including LLaMA, Qwen, Gemma, EuroLLM, and Salamandra, on this benchmark. Our evaluation combines automatic metrics, LLM-as-a-judge approaches (using Prometheus and JudgeLM), and human annotations. Results consistently show superior performance in English, with significantly lower performance in less-resourced languages. While contextual support yields mixed results, it tends to benefit underrepresented languages. These findings underscore the current limitations of LLMs in multilingual commonsense generation.\\
The dataset is publicly available in \url{https://huggingface.co/datasets/gplsi/MULTICOM}.

\end{abstract}

\section{Introduction}
Commonsense reasoning is a fundamental human ability that allows individuals to make inferences without needing everything to be explicitly stated \cite{quyet2024commonsense}. For example, people intuitively know that a dog can bark but not meow, or that a pencil is typically used for writing or drawing. Transferring this type of cognitive skill to Artificial Intelligence (AI) systems is challenging, as it involves enabling models to make inferences beyond what is directly expressed.
% With the emergence of Large Language Models (LLMs), commonsense reasoning in AI has advanced significantly. In particular,
LLMs have become the state of the art in Natural Language Generation (NLG) due to their training on vast amounts of text that capture a wide range of world knowledge. As a result, they are able to exhibit at least a superficial level of commonsense, thanks to the information encoded in their parameters.
However, while LLMs are trained on multilingual data, the majority of that data is in English. This imbalance can lead to disparities in performance across languages~\cite{blasi_systematic_2022,xu_survey_2024,qin_survey_2025}. 
%AG: Some changes below:
% In particular, 
In this paper, we investigate whether LLMs demonstrate the same level of ability to generate text which conforms to commonsense intuitions
% compatible with commonsense, 
across different languages.
% generation ability in underrepresented languages as they do in English.
% Based on this observation, we hypothesise that LLMs do not exhibit consistent commonsense generation capabilities across languages. 
To explore this topic, we use
%surround this research in 
a constrained commonsense generation task, using \dataset{}, a novel dataset inspired by the CommonGen benchmark~\cite{lin2019commongen}: given a triplet of concepts, the model must generate a coherent and plausible sentence that connects them. Additionally, we examine whether providing a supporting context in addition to the concept triple improves the model’s ability to produce meaningful outputs. We evaluate the outcomes across four different languages using automatic metrics, LLM-as-judge evaluations, and human judgments. 
% Our objective is to evaluate and highlight potential limitations in multilingual commonsense generation using a broad benchmark of LLMs.
%The rest of the article is structured as:
Our results show a significant disparity across different languages in the extent to which LLMs are able to generate texts compatible with commonsense. 

In summary, our contributions are as follows: (i) We present \dataset, a novel benchmark dataset for commonsense generation, which extends an existing dataset to encompass four different languages (English, Dutch, Spanish and Valencian); (ii) We introduce and study the role of supporting linguistic context as a way to assist the model during the generation to provide more accurate texts and; (iii) We perform a comprehensive evaluation of commonsense generation capabilities across five large language model families, varying in both size and configuration.

\section{Related Work}
% This section aims to review previous efforts in addressing the task of commonsense text generation, as well as examine how multilingualism has been explored within NLG research.

\paragraph{Commonsense Generation}
Commonsense knowledge refers to the information that people typically infer in everyday situations, even when it is not explicitly stated \cite{bhargava2022commonsense}. Incorporating this type of knowledge into NLG systems has been recognised as both important and challenging \cite{wang-etal-2021-retrieval-enhanced}, motivating several
% As a result,  
several benchmarks and shared tasks.
% have been developed to advance the state of the art in this area.
% One such example is the 
The Avicenna task \cite{aghahadi2022avicenna} focuses on generating a conclusion that logically connects two sentences with a syllogistic relationship. Winograd Schemas test models' ability to resolve referential ambiguities which require commonense reasoning to arrive at the most likely interpretation in a given context~\cite{levesque_winograd_nodate}. Other research has concentrated on more constrained generation tasks, such as keyword-to-text. For instance, the SituatedGen dataset \cite{zhang2023situatedgen} involves generating a pair of contrasting sentences based on a set of concepts, including temporal or geographical entities. Similarly, CommonGen \cite{lin2019commongen} and C$^2$Gen \cite{carlsson2022fine} require generating a coherent and plausible sentence that describes an everyday scenario using a given set of keywords. C$^2$Gen also incorporates contextual information to improve the relevance and accuracy of the generated text.
Despite these advances, most research in this area has focused on English. To address this limitation, the COCOTEROS corpus \cite{maestre2024cocoteros} was introduced to support commonsense text generation in Spanish. Based on the same principles as CommonGen and C$^2$Gen, COCOTEROS aims to generate plausible sentences in Spanish given a set of input concepts.
Nevertheless, there is still a need to evaluate commonsense reasoning across different languages, especially with the emergence of large language models (LLMs) capable of multilingual text generation. The present paper addresses this gap by evaluating the multilingual commonsense generation abilities of LLMs.

\paragraph{Multilinguality in LLMs}
Multilinguality is a key feature of LLMs, as they are trained on massive datasets containing texts in multiple languages. This enables them to understand and generate text across a wide range of languages. As a result, LLMs are often able to produce high-quality outputs in various languages. Nevertheless, recent research suggests that LLM capabilities are not uniform across languages, while many languages remain under-represented in training data~\cite{blasi_systematic_2022,xu_survey_2024,qin_survey_2025}. Several studies have assessed the performance of LLMs across different languages and tasks. For example, \citet{zhang-etal-2023-dont} examine ChatGPT’s performance on three NLP tasks and find that it shows a strong bias toward English, which can hinder its effectiveness in other languages. Similarly, \citet{xuan2025mmlu} introduce MMLU-ProX, a multilingual benchmark derived from English, and report significant performance disparities across languages. \citet{thellmann2024towards} also contribute to this line of research by proposing a benchmark based on translated datasets covering several European languages. In the specific case of constrained, commonsense generation tasks, there has been very limited cross-lingual evaluation, in part due to the paucity of resources. This motivates our development of \dataset{}.

\section{The \dataset{} dataset}
We introduce \dataset{}, a new benchmark to test the capabilities of generative models in commonsense generation tasks. Similarly to CommonGen~\cite{lin2019commongen}, we frame the task in terms of generating text from a triple of keywords or concepts. However, \dataset{} further incorporates contextual support, and crucially covers multiple languages (English, Spanish, Dutch and Valencian), including low-resourced languages.
% Within this section we will explain the benchmark carried out for our experimentation. 

\subsection{Data}
\label{sec:dataset}
% To properly evaluate the abilities of LLMs, it is important to use a dataset that supports the analysis of the two main aspects we focus on: multilingual understanding and contextual helpfulness. For this purpose, 
As our starting point, we use the COCOTEROS dataset \cite{maestre2024cocoteros}, which was specifically created to assess commonsense generation in Spanish. Each instance in the COCOTEROS dataset includes three components:
\begin{itemize}[itemindent=0cm,leftmargin=0.3cm]
\item \textbf{Keywords:} a set of three words that should appear in a generated sentence.
\item \textbf{Context:} a group of sentences that provide semantic or thematic background to support the inclusion of the keywords.
\item \textbf{Target sentence:} a grammatically correct and commonsense-consistent sentence that includes all three keywords or suitable variations.
\end{itemize}

% The dataset is divided into a training set with $3880$ examples and a test set with $969$ examples.
\dataset{} extends COCOTEROS by adding two new elements to each sample:
\begin{itemize}[itemindent=0cm,leftmargin=0.3cm]
\item \textbf{Counterfactual sentence:} a sentence that includes all the keywords and is grammatically correct and fluent in Spanish, but lacks commonsense logic. Counterfactuals were generated using the LLaMA 3.2 model \cite{grattafiori2024llama}, with prompts designed to ensure the inclusion of all keywords, fluency, and the intentional absence of commonsense reasoning. All generated sentences were manually reviewed by the authors to confirm they matched these criteria.
\item \textbf{Unrelated sentence:} a sentence that is grammatically and semantically correct but does not contain any of the keywords for the given instance. These were created by randomly selecting target sentences from other examples in the dataset, making sure there was no overlap with the current instance’s keywords.
\end{itemize}

Following the augmentation of the Spanish data with the new sentence types, we further translated it into three additional languages:

\begin{itemize}[itemindent=0cm,leftmargin=0.3cm]
\item \textbf{English:} We included English based on the hypothesis that LLMs tend to perform better at generating commonsense text in English compared to other languages. We translated the COCOTEROS dataset into English using the OPUS-MT-ES-EN machine translation model~ \cite{TiedemannThottingal:EAMT2020}.\footnote{Model available at \url{https://huggingface.co/Helsinki-NLP/opus-mt-es-en}}.
\item \textbf{Dutch:} Dutch was selected as a medium-resource language, which is somewhat less represented in NLP than English or Spanish.
% to evaluate the performance of LLMs in a less-represented linguistic context than English or Spanish. 
The Dutch version was produced by translating the English subset using the OPUS-MT-EN-NL model~\cite{TiedemannThottingal:EAMT2020}.\footnote{Model available at \url{https://huggingface.co/Helsinki-NLP/opus-mt-en-nl}}
\item \textbf{Valencian:} To explore LLM performance in a low-resource setting, we included Valencian, a regional variety of Catalan spoken in Spain’s Valencian Community. The dataset was translated from Spanish to Valencian using the Salt translation tool provided by the regional government of Valencia \cite{gvaTraductorGeneralitat}.
\end{itemize}

After completing the translations, we used the Grok model \cite{grok2025} to verify that all target sentences in each language contained the intended keywords. This step was necessary due to potential polysemy and translation variation, where a keyword might be rendered differently in context. To address mismatches between keywords and their occurrences in translated sentences, we applied a post-processing step using Grok to align and correct the keyword usage accordingly. The prompt we used within this step is shown in Appendix~\ref{appen:aligmnentgrok}.

The final \dataset{} dataset consists of a training set with $3875$ unique input triples and corresponding context sentences, with references in 4 languages (total: $15,500$) and a test set with $3,876$ instances ($969$ inputs in four languages). Examples from \dataset{} are shown in Appendix~\ref{appen:dataset-examples}.

\subsection{Models}
\label{subsec:models}
We evaluated the commonsense generation capabilities of LLMs across two primary dimensions: model size and instruction tuning (i.e., base vs. instructed versions). To this end, we selected five recent LLM families: LLaMA \cite{grattafiori2024llama}, Qwen \cite{qwen3technicalreport}, Gemma \cite{gemmateam2024gemma2improvingopen}, EuroLLM \cite{MARTINS202553}, and Salamandra \cite{gonzalezagirre2025salamandratechnicalreport}. LLaMA, Qwen and Gemma are widely-used and highly performant LLMs. The rationale for including EuroLLM and Salamandra is their additional focus on multilingual capabilities, through training data curation: EuroLLM is trained on data from all the official languages of the European Union, including Spanish and Dutch. Salamandra was similarly developed with a focus on multilinguality during the data curation process, and further emphasises the languages of Spain, notably Catalan and its varieties, including Valencian.

For each model family, we chose the most recent available models in small and medium sizes, testing both their base and instruction-tuned variants\footnote{Due to compatibility issues, we used version LLaMA 3.2 for the LLaMA family.} for each language. The specific configurations evaluated are as follows:

\begin{itemize}[itemindent=0cm,leftmargin=0.3cm]
\item \textbf{LLaMA}: LLaMA-3.2 in 1B and 3B sizes, including both base and instruct versions (1B-Instruct and 3B-Instruct).
\item \textbf{Gemma}: Gemma-2 3B, tested in both base and instruct variants. In our experiments, we also included Gemmma-9B. However, we observed a strong tendency for the model to misunderstand the task, frequently generating outputs in the form of (pseudo) code. We therefore excluded Gemma-9B from further consideration.
\item \textbf{Qwen}: Qwen3 in 4B and 8B sizes, with corresponding base and instruct versions.
\item \textbf{Salamandra}: Models in 3B and 7B sizes, tested in both base and instructed forms.
\item \textbf{EuroLLM}: EuroLLM models in 1.7B and 9B sizes, evaluated in both base and instructed versions.
\end{itemize}

\subsection{Evaluation Methodology}
\label{subsec:eval}
The evaluation was conducted using three complementary approaches. First, we applied widely-used automatic evaluation metrics to assess model performance quantitatively. Second, we employed LLM-as-a-judge to provide ratings with explicit feedback. Third, a human evaluation was performed on a subset of the dataset to verify the alignment between automated assessments and human judgments.

\paragraph{Automatic metrics} To evaluate the quality of the generated sentences, we employed a range of automatic metrics, comparing generated outputs against both the target and counterfactual sentences in the test set:

\begin{itemize}[itemindent=0cm,leftmargin=0.3cm]
\item \textbf{BERTScore \cite{Zhang2020BERTScore}}: This metric assesses the similarity between a candidate and reference sentence using contextual embeddings from BERT. Unlike traditional approaches, BERTScore captures semantic similarity by computing token-level cosine similarity in the embedding space. To compute BERTScore across all languages in this study, we used the \textit{bert-base-multilingual-cased} model, as its training data includes English, Dutch and Spanish, as well as Catalan, of which Valencian is a variety.
%\item \textbf{Natural Language Inference (NLI)}: This approach assesses whether the generated sentence is more logically entailed by the target or the counterfactual sentence. It provides insight into whether the model's output aligns more closely with the intended or alternative context.
\item \textbf{Universal Sentence Encoder (USE) \cite{cer2018universal} and Cosine Similarity \cite{steck2024cosine}}: To measure semantic similarity, we encode each sentence using the Universal Sentence Encoder to obtain vector representations. Cosine similarity is then computed between these vectors to assess their semantic closeness.
\item \textbf{Dependency Parsing \cite{Jurafsky2009dependency} and Levenshtein Distance \cite{miller2009levenhstein}}:
To evaluate syntactic similarity, we extract dependency triplets (head, relation, dependent) from each sentence using the Stanza \cite{qi2020stanza} pipeline and compare them using symbolic Levenshtein distance.\footnote{We process sentences using the Stanza pipeline configured with the \textit{``tokenize, PoS, lemma, and depparse''} processors. This ensures consistent preprocessing across languages and enables the extraction of universal dependency representations.} Each dependency triplet captures a syntactic relation between a head and a dependent word, labelled with the grammatical relation type. Then, we compare them using the symbolic Levenshtein distance. For each triplet, we compute the average distance across its components and consider it a match if this average falls below a predefined threshold. The final score reflects the proportion of matched triplets relative to the larger set.
\item \textbf{Dependency Parsing \cite{Jurafsky2009dependency} with Vector Representations and Cosine Similarity \cite{steck2024cosine}}: We also assess semantic similarity at the syntactic level. Dependency triplets are extracted using the Stanza dependency parser \cite{qi2020stanza}, and each triplet, comprising a head, relation, and dependent, is converted into a vector by averaging the embeddings of its components. For this conversion, we used SpaCy.\footnote{The models used were: \textit{``en\_core\_web\_md''} for English, \textit{``es\_core\_news\_md''} for Spanish, \textit{``nl\_core\_news\_md''} for Dutch, and \textit{``ca\_core\_news\_md''} for Valencian, due to its close linguistic similarity with Catalan.} Cosine similarity is then computed between vectors from different sentences, and a match is recorded when the similarity exceeds a predefined threshold. The final score is calculated as the proportion of matched triplets relative to the larger set, capturing both semantic content and syntactic structure.
\end{itemize}

\noindent
While the first two metrics above focus on semantic comparison, the metrics relying on dependency triples provide an indication of the extent to which a generated and a reference sentence incorporate the concepts in the input triple in the same or similar syntactic dependency relations. 

\paragraph{LLM-as-Judge} To complement the automatic metrics, we also employed two LLMs-as-a-judge models to assess the commonsense quality of the generated sentences:

\begin{itemize}[itemindent=0cm,leftmargin=0.3cm]
\item \textbf{Prometheus-V2.0} \cite{kim-etal-2024-prometheus}: Prometheus is an LLM specifically designed for evaluating the outputs of other models through both direct scoring and pairwise ranking. It demonstrates strong performance across evaluation formats, approaching the reliability of proprietary models like GPT-4. Prometheus 2 is trained on the preference collection, a comprehensive dataset containing over 1,000 fine-grained evaluation criteria. These criteria go beyond general categories such as helpfulness or harmlessness to include specific skills like commonsense reasoning, making it particularly well-suited for our task. We used Prometheus to assign a score from 1 to 5 based on the rubric shown in Appendix~\ref{appen:rubrics}.

\item \textbf{JudgeLM \cite{zhu2025judgelm}}: JudgeLM is a fine-tuned LLM trained on GPT-4-labelled data to efficiently and reliably evaluate the outputs of other language models, with measures taken during finetuning to support a variety of evaluation formats, and to overcome known biases in LLM-based judgments, such as an excessive reliance on the order in which items are presented. 
% It addresses common evaluation biases and supports a variety of evaluation formats. 
JudgeLM has shown over 90\% agreement with GPT-4, making it a strong proxy evaluator. In our case, JudgeLM provided a score from 0 to 10, based solely on the level of commonsense reasoning demonstrated by each generated sentence. This choice of scale mirrored the scale used for direct assessment during the finetuning process of the model \cite{zhu2025judgelm}. The final prompt used to evaluate the sentences can be seen on Appendix~\ref{appen:judgelm}. To facilitate comparison to Prometheus-2, we normalised the scores returned by JudgeLM to a scale from 1 to 5. 

\end{itemize}

\paragraph{Human evaluation}
For the human evaluation, we selected a random subset of 20 instances from the test set, and collected annotations using the Prolific platform\footnote{Platform available at \url{https://www.prolific.com/}} for each of the output languages \cite{profilic}. For this evaluation, we focused on the outputs of one specific model, LLaMA-3.2-3B-Instruct, which was among the top-performing models according to the results of the automatic metrics and LLM-as-judge evaluation (see Section~\ref{sec:experiments} below).
For each of the target languages in \dataset{}, three native or fluent speakers were selected via the platform.\footnote{No speakers of Valencian are available on Prolific. However, for the purposes of this study, we selected speakers of Catalan, since Valencian is a variety of Catalan. Nevertheless, we urge caution in the interpretation of results for Valencian.} 

\noindent
Annotators were asked to rate the commonsensical quality of generated texts, given the input triple. The instructions mirrored the evaluation rubric used for Prometheus-2 (see Appendix~\ref{appen:rubrics}).

\section{Experiments}
\label{sec:experiments}
%%AG: SUGGESTED STRCUTURE FOR THIS SECTION
%1. Start with a description of the METHOD: we generate both with and without the context sentence. This is crucial to mention
%2. Present the automatic metric results: keep the main details int he appendices. In this section, we should start with an overview of the results, and show only the results for one of the model families, as a representative. E.g. we can just focus on LLaMA. Expand Table 1 by adding rows for all the metrics for LLaMA.
In this section, we compare different models with the goal of evaluating to what extent their commonsense generation capabilities vary across languages. We tested models by prompting them with the input triple and an instruction to generate a sentence using all the words in the triplet. We further compare two different settings, one with the information of a linguistic context included in the prompt (cf. Section~\ref{sec:dataset}) and one without. See Appendix~\ref{appen:prompt} for details of the prompt templates used. 
% This section presents the results obtained from the experimentation. The goal is not to determine which model performs better, but rather to examine whether the models are capable of generating commonsensical sentences across different languages.

\begin{table*}[]
%\scriptsize
\centering
\resizebox{\textwidth}{!}{%
\begin{tabular}{|l|cccc|cccc|cccc|cccc|}
\hline
\multirow{2}{*}{} & 
\multicolumn{4}{c|}{\textbf{ES}} & 
\multicolumn{4}{c|}{\textbf{EN}} & 
\multicolumn{4}{c|}{\textbf{CA-VA}} & 
\multicolumn{4}{c|}{\textbf{NL}} \\ \cline{2-17}
 & Ref. & Count. & Ref. +Con & Count. +Con & Ref. & Count. & Ref. +Con & Count. +Con & Ref. & Count. & Ref. +Con & Count. +Con & Ref. & Count. & Ref. +Con & Count. +Con \\ \hline\hline
\emph{BERTScore | Base} &
0.771 & 0.749 & 0.754 & 0.741 &
0.903 & 0.879 & 0.902 & 0.880 &
0.759 & 0.718 & 0.805 & 0.718 &
0.761 & 0.716 & 0.742 & 0.714 \\ \hline
\emph{BERTScore | Instruct} &
0.777 & 0.754 & 0.743 & 0.727 &
0.908 & 0.881 & 0.902 & 0.876 &
0.765 & 0.725 & 0.756 & 0.700 &
0.771 & 0.722 & 0.746 & 0.707 \\ \hline\hline

\emph{USE | Base} &
0.632 & 0.596 & 0.604 & 0.608 &
0.555 & 0.361 & 0.516 & 0.390 &
0.420 & 0.322 & 0.533 & 0.343 &
0.637 & 0.599 & 0.641 & 0.643 \\ \hline

\emph{USE | Instruct} &
0.650 & 0.619 & 0.549 & 0.538 &
0.558 & 0.364 & 0.509 & 0.360 &
0.440 & 0.347 & 0.422 & 0.304 &
0.672 & 0.632 & 0.639 & 0.623 \\ \hline\hline

\emph{Dep.Par + Lev. | Base} &
0.666 & 0.549 & 0.535 & 0.548 &
0.705 & 0.493 & 0.598 & 0.507 &
0.671 & 0.516 & 0.684 & 0.549 &
0.596 & 0.412 & 0.478 & 0.415 \\ \hline

\emph{Dep.Par + Lev. | Instruct} &
0.680 & 0.577 & 0.508 & 0.502 &
0.718 & 0.525 & 0.621 & 0.506 &
0.693 & 0.554 & 0.576 & 0.499 &
0.621 & 0.436 & 0.492 & 0.405 \\ \hline\hline

\emph{Dep.Par + Cos.Sim | Base} &
0.472 & 0.400 & 0.423 & 0.428 &
0.529 & 0.348 & 0.496 & 0.373 &
0.630 & 0.533 & 0.673 &	0.575 &
0.448 & 0.323 & 0.387 & 0.336 \\ \hline

\emph{Dep.Par + Cos.Sim | Instruct} &
0.492 & 0.428 & 0.399 & 0.383 &
0.556 & 0.369 & 0.498 & 0.369 &
0.658 & 0.571 & 0.552 & 0.501 &
0.472 & 0.336 & 0.402 & 0.318 \\ \hline

\end{tabular}%
}
\caption{Results obtained for the LLaMA-3.2-3B models with automatic evaluation metrics for English (EN), Spanish (ES), Valencian (CA-VA), and Dutch (NL). Comparisons are made to the reference (Ref.) sentence and the counterfactual (Count.) sentence, with and without the context sentence ($\pm$Con). Results for the base and instruction-tuned model are shown in separate rows.}
\label{tab:llamaresults}
\end{table*}

\subsection{Automatic Metrics}
We conducted automatic evaluations using the metrics described in Section~\ref{subsec:eval} across all models. However, due to the extensive volume of results, this section presents only the outcomes for a single model, LLaMA-3.2-3B, which consistently demonstrated strong and stable performance. These results are shown in Table~\ref{tab:llamaresults}. Comprehensive results for all evaluated models are provided in Appendix~\ref{appen:llamaresults} for LLaMA, Appendix~\ref{appen:qwenresults} for Qwen, Appendix~\ref{appen:eurollmresults} for EuroLLM, Appendix~\ref{appen:salamandraresults} for Salamandra, and Appendix~\ref{appen:gemmaresults} for Gemma.

For each model and language, we assessed two types of generated sentences: one without context injection, and another with context injected into the prompt to provide a more constrained input. These generated sentences were then evaluated against both the target sentence and the counterfactual sentence. 
The primary objective was to compare the difference in scores for generated sentences when compared against the target and counterfactual.
% is consistently larger in English than in other languages. 
A larger gap implies that the generated sentence aligns more closely with the reference than with the counterfactual, providing some evidence of the degree ofcommonsense generation abilities in a given language, based on similarity to a reference.

Consistent with the results observed for LLaMA (Table~\ref{tab:llamaresults}), all models generally follow the same trend: the differences in similarity scores, when comparing generated outputs to both target and counterfactual sentences, are consistently larger in English than in other languages. The only exception to this pattern is BERTScore, for which the difference in score between reference and counterfactual in English is comparable to, or slightly smaller than, that observed in other languages. We attribute this to the fact that, although models typically perform better in English, BERTScore may not effectively distinguish between commonsensical and non-commonsensical outputs, as it primarily captures token-level semantic similarity rather than deeper commonsense reasoning.

The impact of contextual information on generation is mixed across models and languages. While certain models—particularly LLaMA—demonstrate improved performance when context is provided, especially for under-represented languages such as Valencian and Dutch, other model families, such as EuroLLM, show only modest gains across all languages. In contrast, for some models like the Gemma family, the inclusion of context appears to reduce performance.

\begin{figure*}[!t]
    \centering
    \includegraphics[width=0.8\textwidth]{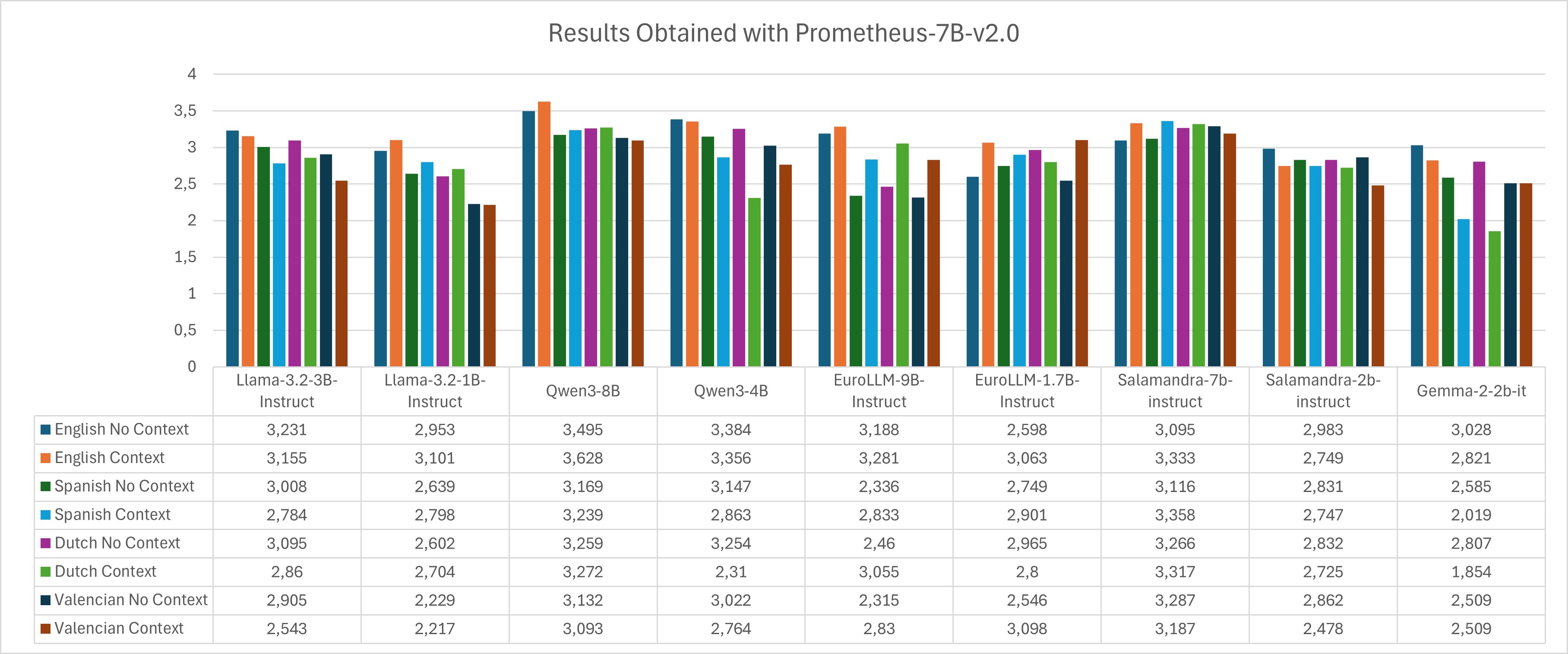}
    \caption{LLM-as-Judge results with Prometheus-2}
    \label{fig:prometheus}
\end{figure*}

\begin{figure*}[!t]
    \centering
    \includegraphics[width=0.8\textwidth]{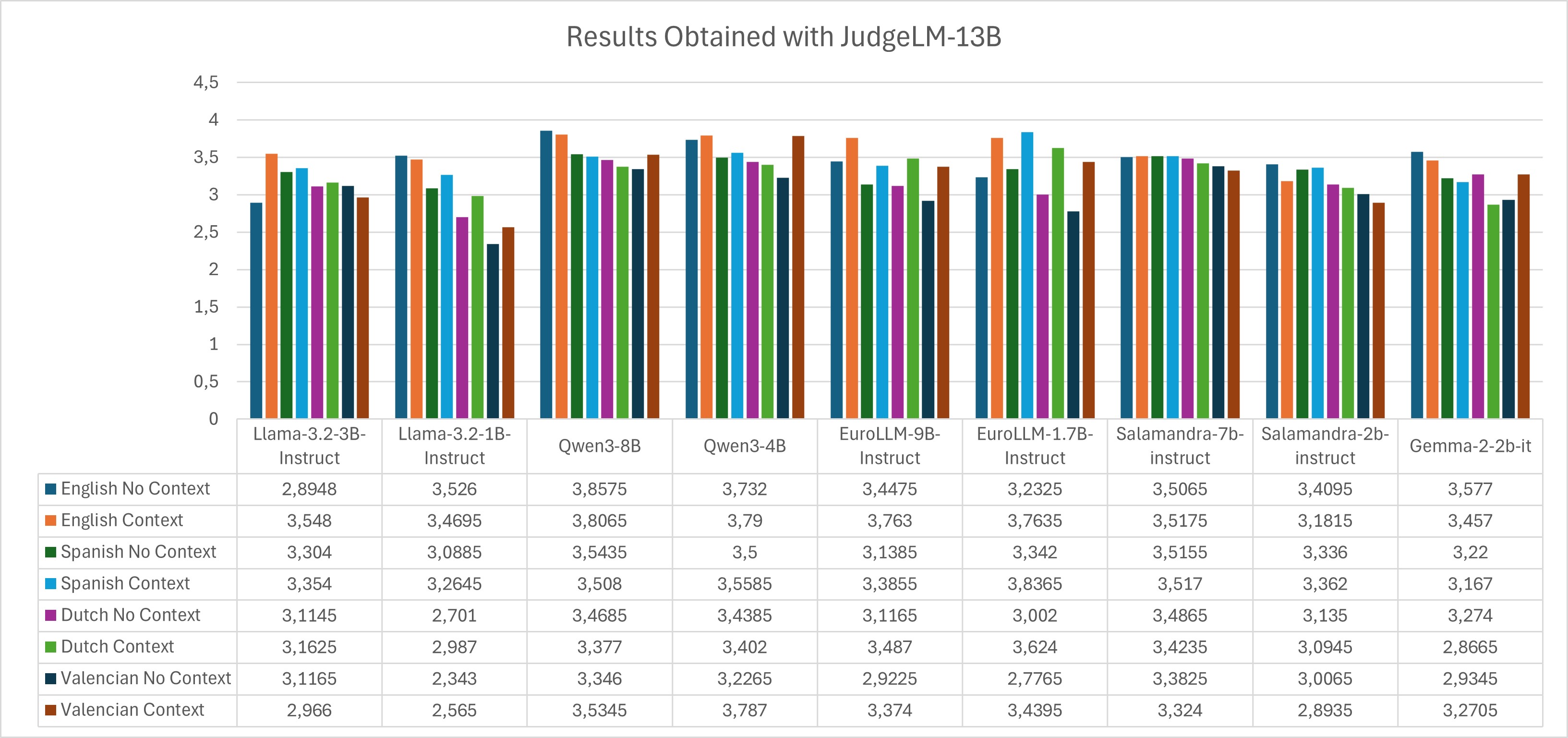}
    \caption{LLM-as-Judge results with JudgeLM, normalised to a 1-5 scale}
    \label{fig:judgelm}
\end{figure*}

\subsection{LLMs as Judges}
For the LLMs-as-a-judge evaluation, we assessed the results of the instruction-tuned versions of each tested model size, as they generally achieved better performance with the automatic metrics compared to their base counterparts. Additionally, we analysed the impact of injecting contextual information during sentence generation. Results from Prometheus-2 are shown in Figure~\ref{fig:prometheus}, while those for JudgeLM are shown in Figure~\ref{fig:judgelm}. 
% As described in Section \ref{subsec:eval}, we employed two models for human-aligned evaluation. The first is Prometheus-2, which assigns a score ranging from 1 to 5, where 5 indicates the highest commonsense and 1 the lowest. The results of our experiments using Prometheus are presented in Figure \ref{fig:prometheus}.

\paragraph{Differences among languages} Consistent with our observations on automatic metrics, the results for LLM-as-judge evaluation suggest a disparity among languages, but this depends on the target models. Broadly speaking, we see a division between models for which the score distribution varies substantially (LLaMA, Qwen, Gemma) across languages, with better scores for English; and models for which the distribution is less obviously skewed towards English (EuroLLM, Salamandra). We attribute this to the nature of the pretraining data for the latter models, where greater attention was paid to linguistic diversity. For LLaMA and Qwen models, which are among the best-performing on the automatic metric evaluation, output in English is judged as superior by both models, albeit with some variation between the two LLM judges on the impact of context (see below). Surprisingly, JudgeLM scores outputs in Valencian in the {\em context} condition on a par with English. On the other hand, the larger version of Salamandra obtains a more even score distribution across languages with both LLM judges.

\paragraph{Impact of context sentence} The injection of a supporting context sentence does not appear to exert a significant impact on results, with two exceptions: JudgeLM scores LLaMA-3B outputs for English, and Qwen-4B outputs for Valencian, significantly higher when context is included. Nevertheless, this tendency is not consistent across the two LLM judges.

\paragraph{Effect of model size} Larger models can be seen to perform somewhat better than smaller ones, though not by a wide margin. However, larger models do tend to be judged better when generating in under-resourced languages like Valencian, with the possible exception of EuroLLM.

\paragraph{LLM judge disparities} Beyond the observations made above, it is worth noting that the two LLM-as-judge models display some inconsistencies even for the same model configurations and languages, suggesting that model-based judgments may depend on the specific fine-tuning regime employed for a specific LLM-as-judge.

\subsection{Human Evaluation}
For the human evaluation, we focused on responses generated by the LLaMA-3.2-3B-Instruct model, as it showed strong performance across the automatic metrics in our experiments. In this case, we evaluated only one model across multiple languages without context. In what follows, we compare the results of the human judgments with those of Prometheus-2 and JudgeLM, for the same sample of 20 instances in the \dataset{} test set.
% aiming to make a direct comparison with the LLM-as-a-judge results and to assess the alignment between human annotations and automated judgments.

\begin{figure}[h!]
    \centering    \includegraphics[width=0.45\textwidth]{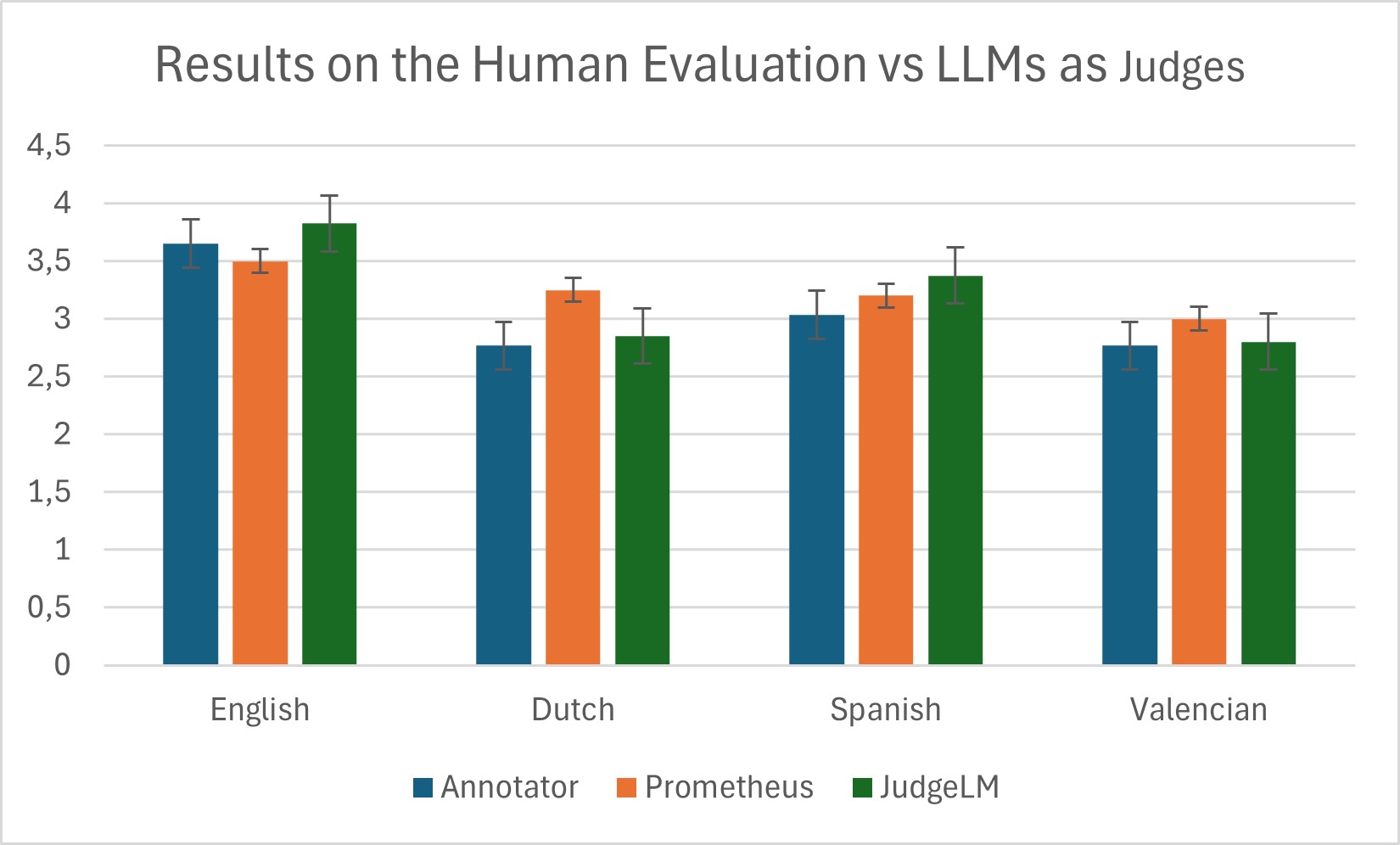}
    \caption{Results obtained for the tested models with the human evaluation.}
    \label{fig:humaneval}
\end{figure}

% Annotators were instructed to rate the commonsense quality of each response on a 1–5 Likert scale, using the same evaluation rubric provided to Prometheus. A subset of 20 sentences generated by LLaMA-3.2-3B-Instruct was selected for evaluation. 
The results of the human annotation process are shown in Figure~\ref{fig:humaneval}.
Based on the results from the human evaluation, we can confirm that human judgments generally align with those of the LLM-as-a-judge evaluators. In particular, the English-generated sentences tend to demonstrate stronger commonsense reasoning compared to other languages. Interestingly, although Spanish is generally considered the second-best language, Prometheus shows a slight preference for Dutch, which it ranks above Spanish as the third-best language overall. In contrast, the other evaluations consistently rank Spanish as the second-best language, while Valencian and Dutch yield comparable results. This discrepancy may be due to Prometheus having been trained on very limited data in Dutch, which could impair its ability to accurately evaluate texts in that language. We further note that the annotators for Valencian are speakers of Catalan; hence they may have somewhat more limited knowledge of the Valencian variety.

To assess consistency among the three annotators, we calculated the majority agreement percentage, which reflects the proportion of items where at least two annotators assigned the same label. Given the inherently subjective nature of commonsense evaluation, achieving perfect consensus among all annotators is often unrealistic. Therefore, majority agreement serves as a practical and meaningful measure of reliability in this context. The observed scores—0.75 for English and Dutch, 0.80 for Spanish, and 0.95 for Valencian—indicate a generally strong level of agreement. These values suggest that despite some variability in individual judgments, the annotators largely concur on the commonsense quality of the responses, supporting the validity of the human evaluation results.

\section{Conclusions}
This study provides key insights into the multilingual commonsense generation capabilities of LLMs through evaluation on the \dataset{} benchmark.

\paragraph{English commonsense predominance}
Our findings indicate a clear imbalance in LLMs' ability to generate commonsensical sentences across languages. Although these models are trained on multilingual data and can produce text in several languages, they demonstrate a significantly stronger performance in English. This is because, while LLMs tend to generate commonsensical outputs in English, they often struggle to do so in other languages. This finding aligns with the hypothesis of this paper, and was supported through three distinct evaluation approaches: (1) automatic evaluation using standard metrics, (2) LLM-based evaluation with LLMs as judges, and (3) human evaluation on a small subset of data. 

\paragraph{Context Disparity}
During the experimentation, we designed two main setups for the commonsense generation. In the first, the models were prompted to generate a commonsensical sentence based on a given triplet of keywords. In the second, we included in the prompt a linguistic context intended to support the model in producing a more plausible output. However, the results obtained from the context-enhanced setup were not as expected. Instead of consistently improving the performance, the inclusion of context resulted in mixed outcomes. While some models seem not to benefit from the context, in a few cases it even had a detrimental impact. In contrast, there were some cases where models demonstrated improved performance when enhanced with the context. This was observed with EuroLLM and Salamandra. This variation suggest that the effectiveness of context injection is model dependent, potentially influenced by the model's training or instruction tuning process. Another important observation concerns language dependence. For under-represented languages such as Valencian and Dutch, contextual information generally had a more positive impact compared to major languages like English and Spanish. This may be because pre-trained models already possess richer representations for high-resource languages, and adding external context could potentially introduce noise rather than improve understanding.

\paragraph{LLMs versus human judges}
To evaluate the outputs of our experiments, we leveraged the capabilities of LLMs-as-judges to assess a challenging task such as commonsense generation. 
% Specifically, we employed two of the most used evaluators: Prometheus and JudgeLM. 
Additionally, we conducted a human evaluation via the Prolific platform on a subset of generated sentences, aiming to verify the alignment between human judgments and those produced by the automated judges. The results obtained from Prometheus and JudgeLM were largely consistent, with minor variations. Both models supported the conclusions drawn from automatic metrics, English being the language in which models most reliably generated commonsensical sentences. Similarly, both types of evaluators consistently assigned the lowest scores to outputs in Valencian, confirming that models struggled the most in this underrepresented language.
These trends were further validated by human evaluation. We compared the human-assigned scores with those from Prometheus and JudgeLM on the same set of sentences. The analysis revealed that humans and LLM judges are broadly compatible, particularly in ranking English as the top-performing language by a substantial margin. This alignment reinforces the reliability of LLM-as-a-judge approaches for commonsense evaluation, while also highlighting persistent performance disparities across languages.

\paragraph{Future Work}
Building on the findings of this study, we identify several promising directions for future research aimed at advancing multilingual commonsense evaluation and generation. One possible direction is to further fine-tune an LLM-as-judge model with the goal of improving alignment between model judgments and the nuanced demands of multilingual commonsense reasoning. Second, while this study has focused on open-weight models, it would be useful to compare their performance to proprietary onessuch as GPT-4, Grok, and Gemini. Finally, given the results of this study, an important avenue for future work is to improve the commonsense reasoning and generation capabilities of models in low-resource languages, for example through further training and/or injection of external knowledge.

% \paragraph{Enhance low-resource language performance} by exploring techniques for injecting external knowledge, thereby improving the commonsense reasoning capabilities of LLMs in underrepresented languages.

%\begin{itemize}
%\item Train a dedicated LLM-as-a-judge using our compiled evaluation corpus, with the goal of improving alignment between model judgments and the nuanced demands of multilingual commonsense reasoning.
%\item Compare open-source and proprietary models by investigating differences in commonsense generation across languages, including models such as GPT-4, Grok, and Gemini.
%\item Enhance low-resource language performance by exploring techniques for injecting external knowledge, thereby improving the commonsense reasoning capabilities of LLMs in underrepresented languages.
%\end{itemize}

\section*{Limitations}

This study presents some limitations that may have influenced the evaluation results.

First, while we employed both automatic metrics and LLM-as-a-judge approaches, the automatic evaluation relied on language-specific encoders and parsers. These tools are typically developed and benchmarked primarily on English, which may lead to more favourable performance scores for English texts. Consequently, the evaluation of non-English outputs might not fully reflect their true quality, and could under-represent the commonsense reasoning abilities of models in less-resourced languages.

Second, for the LLM-as-a-judge evaluations, both prompts and scoring rubrics were written in English, even when evaluating outputs in other languages. While this approach ensures consistency across evaluations, it may introduce subtle biases, particularly if the judging LLM has varying levels of exposure to the target language. This could affect the model’s ability to fully capture the nuances of commonsense reasoning in certain languages, potentially influencing the reliability of cross-linguistic comparisons.

Third, our evaluation was limited to open-source models. While proprietary models such as GPT-4, Grok, or Gemini may exhibit different behavioural patterns or performance characteristics, they were excluded from this study due to the significant costs and access restrictions associated with their use. By focusing on open-source models, we aimed to ensure full transparency, reproducibility, and accessibility of our experimental setup, which are essential for fostering open research and facilitating further comparative studies.

Finally, for the evaluation of Valencian, we were restricted by the availability of models and participants. Due to the paucity of resources for this language, we relied on models trained on Catalan for some parts of the study (such as the computation of BERTScore), and on Catalan speakers for the human evaluation. While this means that results for Valencian should be treated with caution, we also hope that its inclusion in \dataset{} will motivate further work on this language.

\section*{Ethical considerations}
For our human evaluation, participants recruited via Prolific were paid at an hourly rate of \textsterling9.20, as recommended by the platform. The median time of completion of the study was 15 minutes.

To our knowledge, the data in \dataset{} does not include any explicitly harmful text. We therefore have no reason to believe that its public release will invite harmful (dual) uses. At the same time, since the data was based on an existing Spanish dataset \cite[COCOTEROS;][]{maestre2024cocoteros}, which was initially sourced from human annotators, we cannot exclude the possibility of cultural or linguistic biases having crept into the data creation process.

\section*{Acknowledgments}
The research work conducted is part of the R\&D projects ``CORTEX: Conscious Text Generation'' (PID2021-123956OB-I00), funded by MCIN/ AEI/10.13039/501100011033/ and by ``ERDF A way of making Europe''. Moreover, the experiments were carried out using the DGX units within the framework of project IDIFEDER/2020/003, co-funded by the Valencia Government and the European Regional Development Fund (ERDF).

\bibliography{custom}

\appendix

\section{Keyword-Sentence Alignment with Grok}
\label{appen:aligmnentgrok}
As explained in Section~\ref{sec:dataset}, during the translation of the corpus into other languages, there were instances where the translated keywords were not present in the corresponding target sentences due to word polysemy. To address this issue, we applied a post-processing step using Grok to ensure proper alignment between keywords and sentences throughout the whole dataset. The prompt used for this task is shown below. In this example, we illustrate the alignment process for the Valencian translation; for other languages, we simply replaced ``Valencian'' with the corresponding language name.\begin{lstlisting}[basicstyle=\footnotesize\ttfamily, breaklines=true, frame=single]

I have a TSV file with a column named keywords, which contains Valencian keywords. These keywords are intended to be translations of the Spanish keywords in the keywords_es column. However, the translations in the keywords column are often incorrect or do not appear in the corresponding reference sentence (written in Valencian).

Your task is to revise the keywords column so that:
1. Each keyword is a correct Valencian translation of the corresponding word in the keywords_es column.
2. Each translated keyword must appear exactly in the associated reference sentence.
3. If a correct translation of a keyword is not present in the reference sentence, omit it from the final list of keywords, and include another word from the sentence.
\end{lstlisting}

\section{Examples from \dataset{}}\label{appen:dataset-examples}

Examples from the MULTICOM dataset are provided in Table \ref{tab:datasetsamples}.

\begin{table*}[]
\centering
\begin{tabular}{|p{1.6cm}|p{3.5cm}|p{2.5cm}|p{3.8cm}|p{2.5cm}|}
\hline
\textbf{Keyword} & \textbf{Context} & \textbf{Target Sentence} & \textbf{Counterfactual Sentence} & \textbf{Unrelated Sentence} \\
\hline
\makecell[tl]{learn,\\piano,\\dad} &
They gave me a musical instrument to develop my talent. &
My dad encouraged me to learn the piano. &
My dad's piano makes me learn to dance on the moon with his plastic shoes. &
Please give me something to eat. \\
\hline
\makecell[tl]{open,\\man,\\door} &
The noise of the lock woke the woman. She looked at the door and saw a dark figure walking out the hall. &
The door opened and a man came out. &
The woman opened the door with an invisible man who danced tango. &
I have to remove a lot of files from my computer. \\
\hline
\makecell[tl]{read,\\book,\\night} &
After a long day’s work, Maria decided to spend the night at home. She didn’t feel like going out, so she made herself a cup of tea and plunged into a new book. &
Tonight I'd rather read a book than go out. &
Mary's night became a night of reading without a book. &
There is so much beauty in ambiguity. \\
\hline
\makecell[tl]{have,\\negotiation,\\come} &
The lack of consensus prevented agreement from being reached. &
We had a lot of negotiations but they didn't come to anything. &
The negotiation of having a cup of coffee reaches the limit of denial. &
Thank you very much for your letter of 7 January. \\
\hline
\end{tabular}
\caption{Samples from the MULTICOM dataset. For readability, only English samples are shown, but they are aligned across languages.}
\label{tab:datasetsamples}
\end{table*}

\section{Rubric Used for Prometheus Evaluation}
\label{appen:rubrics}
To evaluate the generated outputs using Prometheus as an evaluator, it is essential to carefully define the evaluation rubrics that guide the model in assigning a final score to each sentence. The rubrics used for this evaluation are as follows:\begin{lstlisting}[basicstyle=\footnotesize\ttfamily, breaklines=true, frame=single]
RUBRICS = {
"criteria": "Does the model's answer demonstrate commonsense capability?",
"score1_description": "No Commonsense. The answer is illogical, implausible, or nonsensical from a commonsense perspective. It may reflect a fundamental misunderstanding of how the world works.",
"score2_description": "Poor Commonsense. The answer is grammatically readable but lacks commonsense reasoning. It may include unrealistic, contradictory, or illogical actions or relationships.",
"score3_description": "Limited Commonsense. The answer contains a plausible idea, but it may be vague, confusing, or reflect partial misunderstandings of how the world works. The logic is somewhat strained, but the meaning is recoverable.",
"score4_description": "Good Commonsense. The answer makes sense and reflects commonsense understanding, though it may be slightly awkward or overly simplistic. Any issues present do not significantly affect its plausibility.",
"score5_description": "Excellent Commonsense. The answer demonstrates strong commonsense reasoning. It reflects how people typically think and behave in the real world."
}
\end{lstlisting}

\section{Prompt used for JudgeLM Evaluation}
\label{appen:judgelm}
The JudgeLM model is trained to process an instruction provided to the generation models, the corresponding target response, the generated response, and the specific evaluation aspect to focus on, along with the desired scoring template. Based on this framework, we designed the following prompt to guide the evaluation process using JudgeLM:\begin{lstlisting}[basicstyle=\footnotesize\ttfamily, breaklines=true, frame=single]
### Instruction:
Your task is to construct a commonsensical and grammatically correct sentence in {lang}, using all of the given words. The sentence should sound natural and make logical sense within everyday usage. 
### Keywords:
{keywords}
### Response 1: {target_sentence} 
### Response 2: {generated_sentence}
###Evaluation: Evaluate the two answers based on their level of commonsense reasoning. Assign a score from 0 to 10 to each answer, where 0 indicates no commonsense reasoning and 10 indicates excellent commonsense reasoning. For each score, include a brief explanation justifying your evaluation. Format your response exactly as shown below, replacing [Score] with your assigned score and [Explanation] with your justification:
### Answer 1: [Score]/10 - [Explanation]
### Answer 2: [Score]/10 - [Explanation]
\end{lstlisting}

\section{Prompts for Generating the Sentences}
\label{appen:prompt}
To generate the sentences, we designed a descriptive prompt using a one-shot configuration, in which a single example demonstrated the task to be performed. This prompt was identical across all models and written in English for all target languages, as our preliminary analysis showed that models better understood the task when instructions were provided in English. However, the example within the prompt was written in the target language to guide the model appropriately. For readability, we present the example in English here. Additionally, we used a placeholder {lang} to indicate the target language for generation. The prompt used in the configuration without context is as follows:
\begin{lstlisting}[basicstyle=\footnotesize\ttfamily, breaklines=true, frame=single]
Your task is to construct a commonsensical and grammatically correct sentence in {lang}, using all of the given words. The sentence should sound natural and make logical sense within everyday usage. I will provide an example to show you what is expected:

### Keywords:
cat, owner, pet
### Sentence:
The owner pets the cat.

Your turn:
### Keywords:
{example["keywords"]}:
### Sentence:
}
\end{lstlisting}
For the configuration with context injection, the contextual information was incorporated directly into the prompt as follows:
\begin{lstlisting}[basicstyle=\footnotesize\ttfamily, breaklines=true, frame=single]
Your task is to construct a commonsensical and grammatically correct sentence in {lang}, using all of the given words, and adhering to the given context. The sentence should sound natural and make logical sense within everyday usage. I will provide an example to show you what is expected:

### Keywords:
cat, owner, pet
### Context:
The person arrived home after work.
### Sentence:
The owner pets the cat.

Your turn:
### Keywords:
{example["keywords"]}
### Context:
{example["context"]}
### Sentence:
}
\end{lstlisting}

\section{Results with LLaMA models}
\label{appen:llamaresults}

\begin{figure*}[h!]
    \centering
    \includegraphics[width=0.9\textwidth]{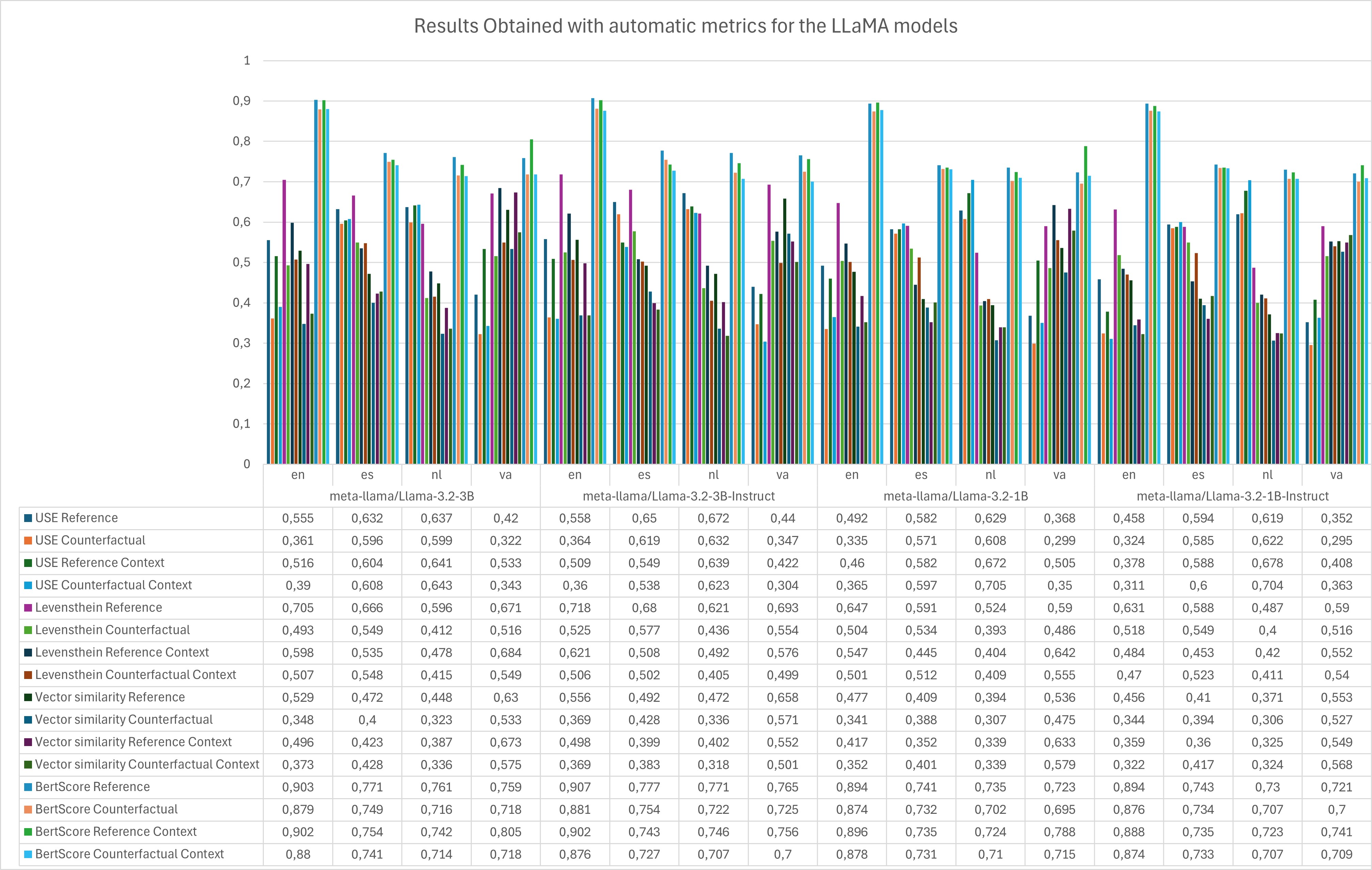}
    \caption{Results obtained for the LLaMA models with the automatic metrics.}
    \label{fig:llamaresults}
\end{figure*}

Regarding the results obtained for the LLaMA models, we can see them in Figure~\ref{fig:llamaresults}

The goal is to demonstrate that, for each language, the difference in similarity scores between the generated sentence and the reference sentence versus the generated sentence and the counterfactual sentence (e.g., using the USE evaluation method) is greater in English than in other languages. A larger difference indicates that the generated sentence is more similar to the reference and less similar to the counterfactual, suggesting stronger commonsense reasoning.

We analyse three key aspects:

\paragraph{Differences among languages} 
In the case of LLaMA models, a consistent trend is observed: the difference in similarity scores between the generated sentence and the reference versus the counterfactual is substantially larger in English than in other languages, where this gap is notably smaller. This suggests that the models perform better at commonsense reasoning in English.

\paragraph{Impact of context sentence}
The role of context in sentence generation appears ambiguous. For high-resource languages (English and Spanish), results show that adding context does not lead to significant improvements. However, in lower-resource languages (Dutch and Valencian), context seems to help produce more commonsensical outputs. This benefit is particularly evident in the base versions of the models, where outputs generated with context obtain higher similarity scores compared to those without.

\paragraph{Effect of model size}
Larger models consistently achieve slightly higher scores across all languages. This is expected, as larger models have been trained on more extensive multilingual data, enhancing their ability to generate sentences that align better with commonsense knowledge.

\section{Results with Qwen models}
\label{appen:qwenresults}
\begin{figure*}[h!]
    \centering
    \includegraphics[width=0.9\textwidth]{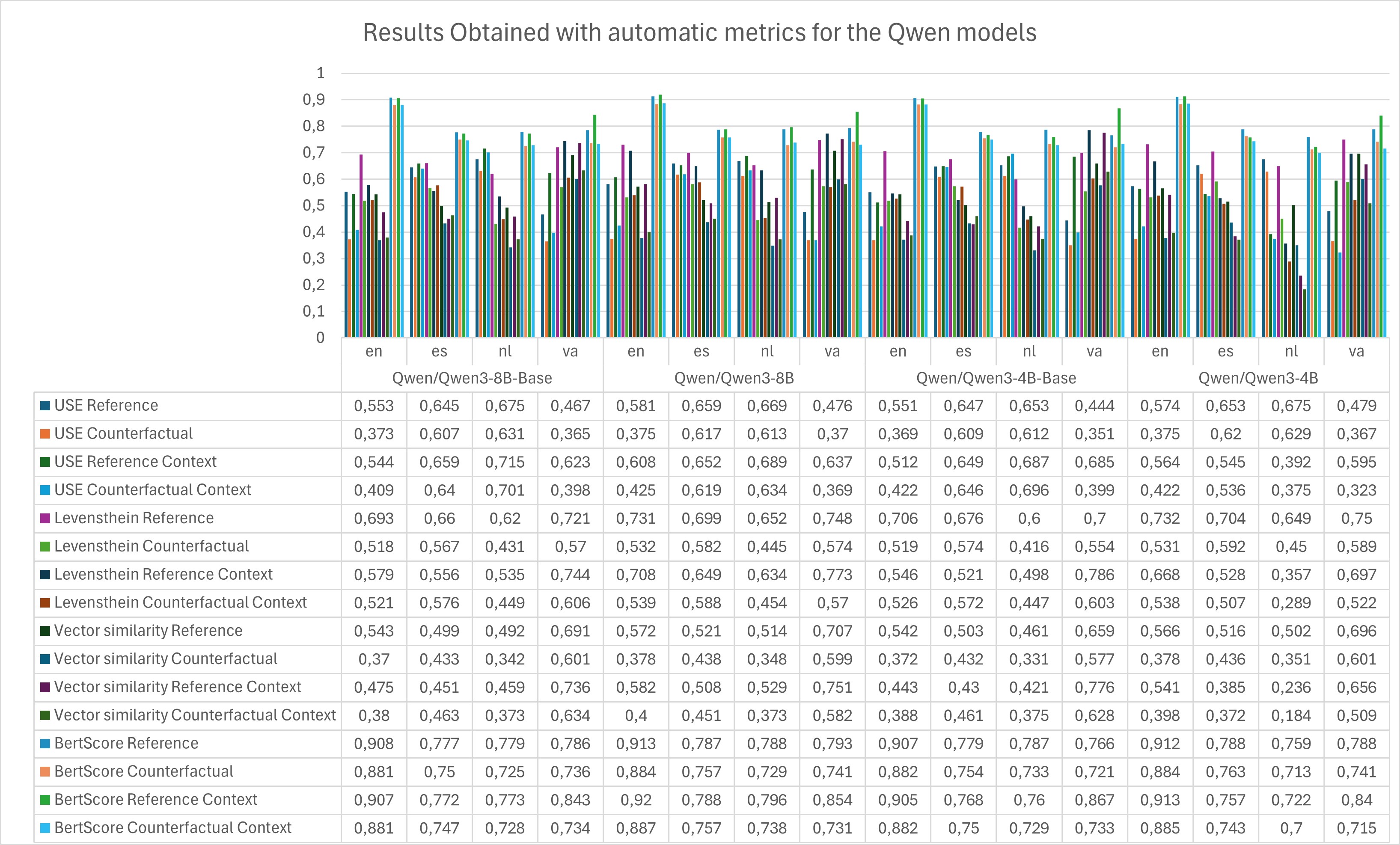}
    \caption{Results obtained for the Qwen models with the automatic metrics.}
    \label{fig:qwenresults}
\end{figure*}

The results obtained by the Qwen models are shown in Figure~\ref{fig:qwenresults}. These models follow a similar trend to those observed with the LLaMA models.

\paragraph{Differences among languages} 
The largest differences between the reference and counterfactual comparisons are observed in English, indicating stronger commonsense reasoning capabilities in that language. In contrast, the differences are minimal for Spanish and Dutch, suggesting that the generated sentences in these languages exhibit limited commonsense reasoning. Interestingly, Valencian shows a greater difference than Spanish and Dutch. This unexpected result may be attributed to limitations in the sentence encoders used for Valencian, which may fail to accurately capture the nuances of the language.
A notable observation for the Qwen models is that, when evaluated using Levenshtein distance, the differences remain relatively high across all languages.

\paragraph{Impact of context sentence}
Another key finding relates to the impact of contextual information. Overall, providing additional context appears to have a limited positive effect on performance. However, in less-represented languages, such as Valencian and Dutch, the inclusion of context tends to yield better results. This suggests that context injection may help compensate for the models’ weaker prior knowledge in these languages, leading to more commonsensical outputs.

\paragraph{Effect of model size}
With respect to model size, there appears to be no significant impact on the commonsensical quality of the generated sentences. The scores obtained using automatic evaluation metrics are similar across different model sizes, suggesting that increasing model size does not necessarily lead to better commonsense reasoning in this context.

\section{Results with EuroLLM models}
\label{appen:eurollmresults}
\begin{figure*}[h!]
    \centering
    \includegraphics[width=0.9\textwidth]{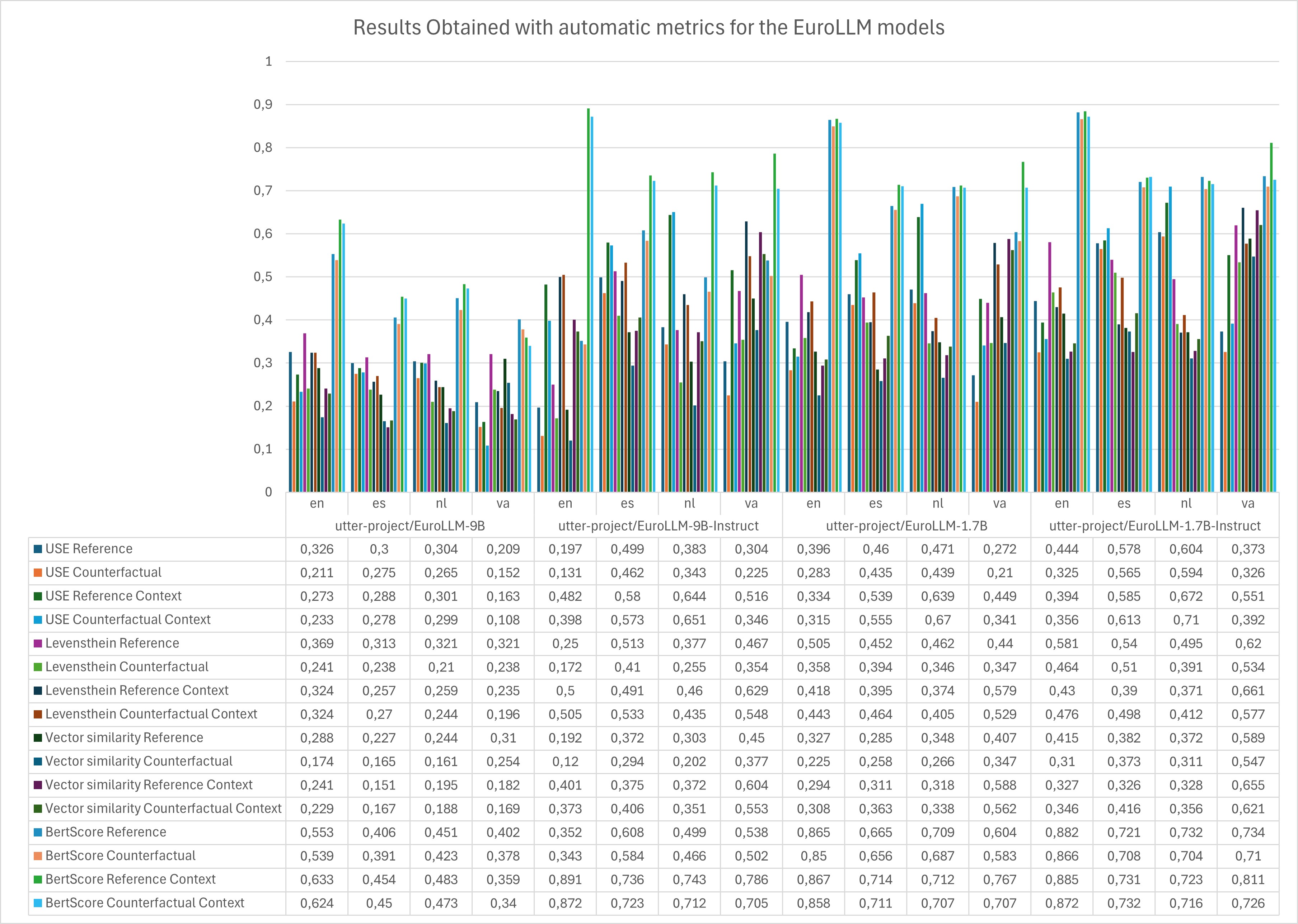}
    \caption{Results obtained for the EuroLLM models with the automatic metrics.}
    \label{fig:eurollmresults}
\end{figure*}

In the case of the EuroLLM models, the results are presented in Figure~\ref{fig:eurollmresults}.

\paragraph{Differences among languages} 
These models exhibit a better ability to generate commonsensical sentences in English compared to other languages. This is reflected in the larger gap between the similarity scores of the generated sentence when compared to the reference versus the counterfactual. Notably, for some non-English languages, the generated sentence is, in fact, more similar to the counterfactual than to the reference. This suggests that the EuroLLM models are more prone to generating nonsensical outputs relative to the previously discussed models. As well as for Qwen, is also observed a bigger difference for the Valencian, that can be due the use of the encoder, which may be not working well. 

\paragraph{Impact of context sentence}
In the EuroLLM model family, context appears to have an overall positive impact. Most models benefit from the inclusion of contextual information, consistently showing improved results across languages. The only exception is the larger variant of the base model, where the addition of context does not provide the same level of improvement observed in the other models.

\paragraph{Effect of model size}
Curiously, within the EuroLLM family, the smaller model versions consistently yield better results across all languages. This may be attributed to differences in training data exposure, as larger models may have been exposed to more diverse or noisier data, which could hinder their ability to generate consistently commonsensical outputs across languages.

\section{Results with Salamandra models}
\label{appen:salamandraresults}
\begin{figure*}[h!]
    \centering
    \includegraphics[width=0.9\textwidth]{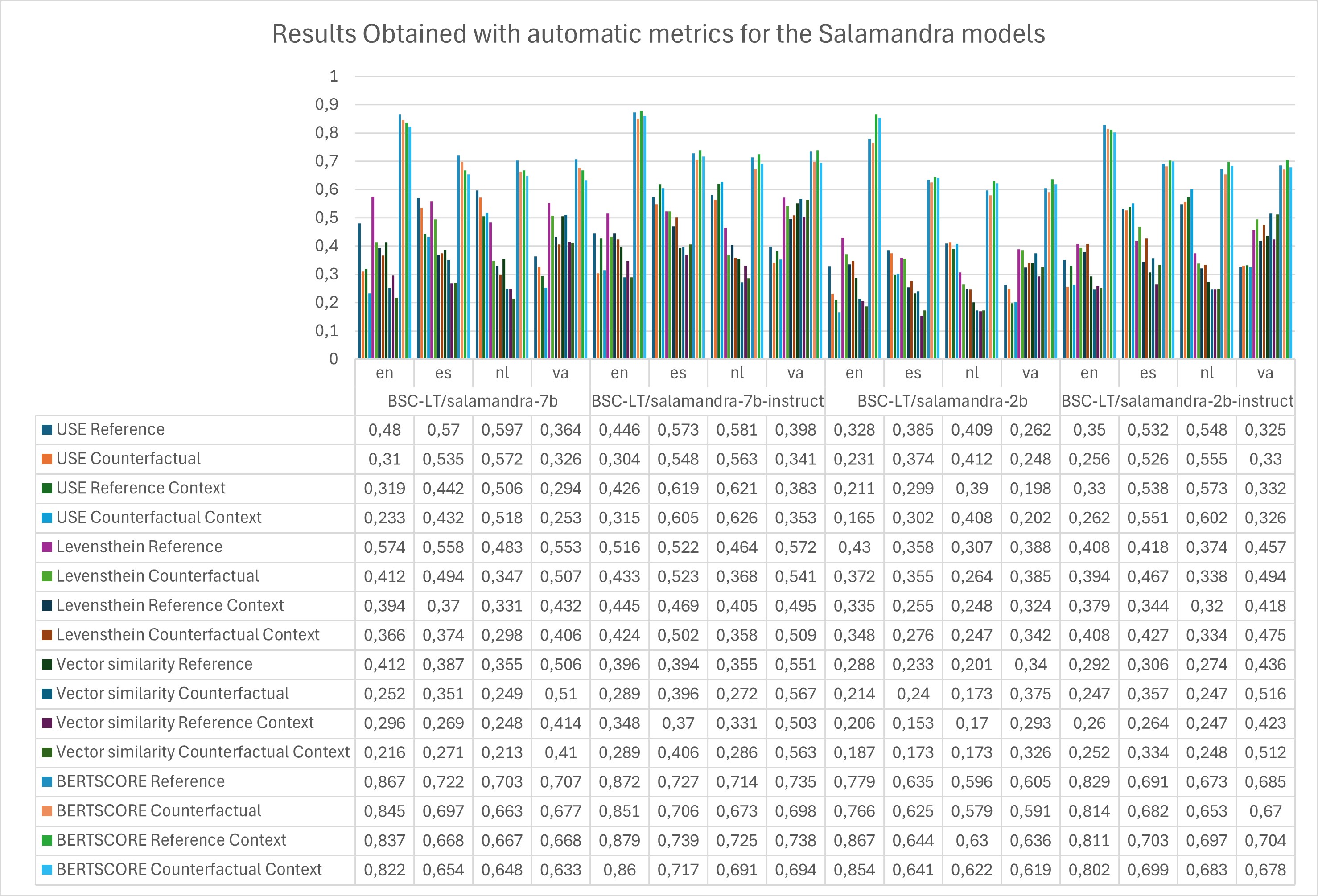}
    \caption{Results obtained for the Salamandra models with the automatic metrics.}
    \label{fig:salamandraresults}
\end{figure*}

The results for the Salamandra models are presented in Figure~\ref{fig:salamandraresults}.

\paragraph{Differences among languages} 
 These results show that English continues to exhibit the largest differences in similarity scores when comparing the generated outputs against the reference and counterfactual sentences. This further supports the observation that, for English, models are more likely to generate commonsensical responses. Notably, this model family was trained on extensive data in Catalan, a variant of Valencian. However, the scores showed in the automatic metrics does not shows as much improvement, as the results for other languages shows a bigger distance. 
 
\paragraph{Impact of context sentence}
In the case of Salamandra models, adding context provides a slight improvement in the instructed versions. However, for the base versions, the inclusion of context appears to actually lower the performance, indicating that context may sometimes hinder rather than help the generation process in these models.

\paragraph{Effect of model size}
Regarding commonsense capability and model size, as expected, larger models demonstrate a better ability to generate commonsensical sentences. This is likely because they have been trained on larger volumes of text, allowing them to learn and reproduce commonsense knowledge more effectively.

\section{Results with Gemma models}
\label{appen:gemmaresults}
\begin{figure*}[h!]
    \centering
    \includegraphics[width=0.9\textwidth]{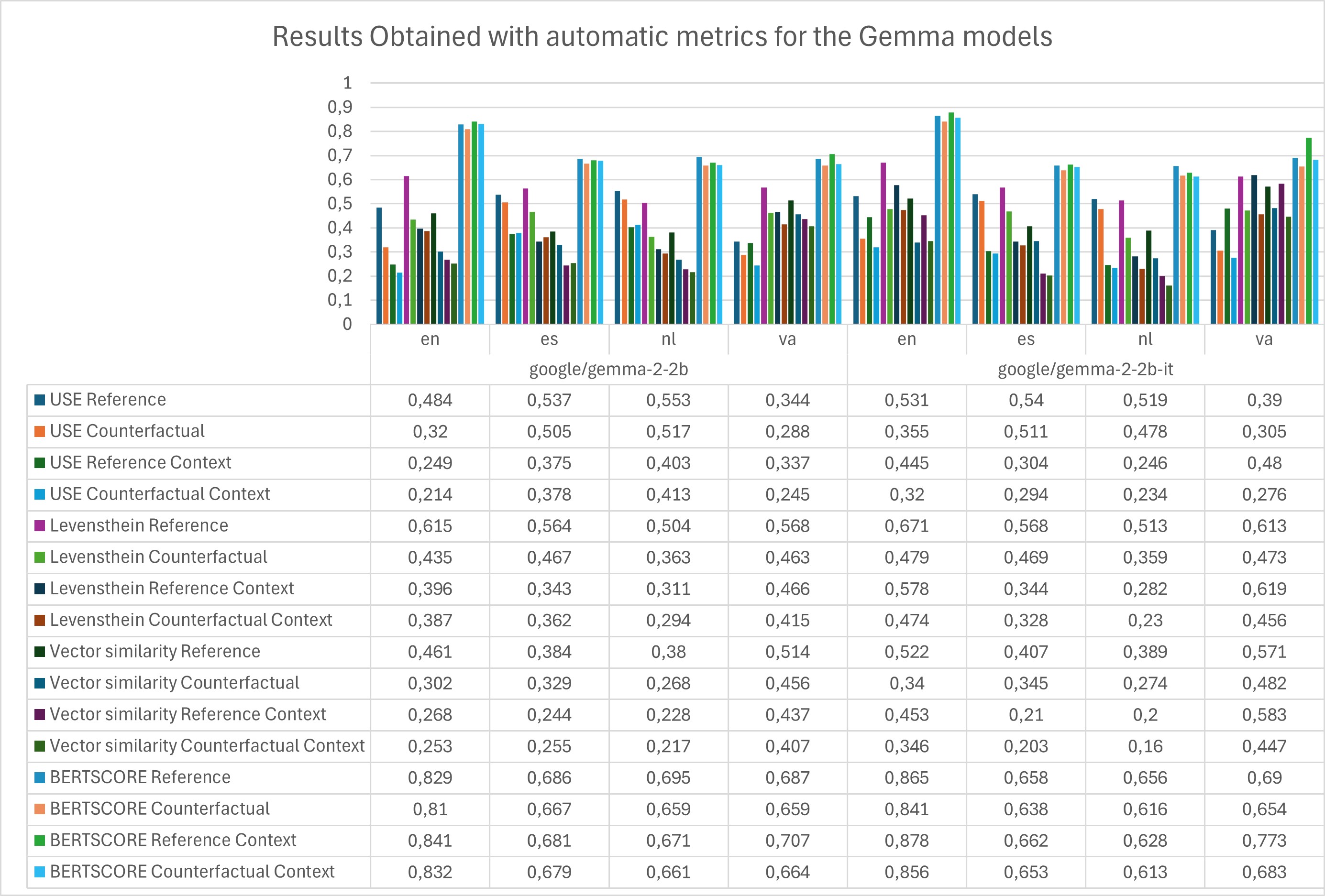}
    \caption{Results obtained for the Gemma models with the automatic metrics.}
    \label{fig:gemmaresults}
\end{figure*}

The results from evaluating the Gemma models are shown in Figure~\ref{fig:gemmaresults}.

\paragraph{Differences Among Languages}
For this model, the difference in similarity scores is notably higher in English than in other languages, reflecting stronger commonsense reasoning capabilities in English. In contrast, the results for Valencian reveal an inverse pattern: the generated sentences are more similar to the counterfactual than to the reference, suggesting a lack of commonsensical understanding. This may be attributed to the model’s limited or nonexistent exposure to training data in Valencian.

\paragraph{Impact of Context Sentence}
Regarding contexts, these models do not seem to benefit from their inclusion. In fact, the scores obtained when context is injected are often lower than those without context, indicating that adding context may sometimes hinder performance.

\end{document}